\documentclass[runningheads]{llncs}
\renewcommand{\inst}[1]{\leavevmode\unskip$^{#1}$}
 
\usepackage[mobile]{eccv}

\usepackage{eccvabbrv}

\usepackage{graphicx}
\usepackage{booktabs}
\usepackage{xspace}

\usepackage[accsupp]{axessibility}  

\usepackage{hyperref}

\usepackage{orcidlink}

\usepackage{graphicx}
\usepackage{multirow}
\usepackage{pifont}
\usepackage{algorithm}
\usepackage{algpseudocode}
\usepackage[normalem]{ulem}
\usepackage{amsmath}
\usepackage{wrapfig}
\usepackage{colortbl}
\usepackage{cuted}
\usepackage{subcaption}
\usepackage{tabularx}
\usepackage{placeins}

\begin{document}

\title{Honey, I Shrunk the Arc de Triomphe!} 


\newcommand{\misscite}{\textcolor{red}{[C]}}
\newcommand{\missref}{\textcolor{red}{[R]}}
\providecommand{\eg}{\textit{e.g.}\@\xspace}
\providecommand{\ie}{\textit{i.e.}\@\xspace}
\providecommand{\wrt}{\textit{w.r.t.}\@\xspace}
\providecommand{\etal}{\textit{et al}\@\xspace}

\newcommand{\amber}[1]{\textcolor{orange}{{[#1]}}}

\newcommand{\metricscenes}{MetricScenes\xspace}

\definecolor{darklavender}{rgb}{0.45, 0.31, 0.59}
\definecolor{darkviolet}{rgb}{0.58, 0.0, 0.83}

\definecolor{visible-blue}{rgb}{0.286, 0.525, 0.910}

\definecolor{tabfirst}{rgb}{1, 0.7, 0.7} 
\definecolor{tabsecond}{rgb}{1, 0.85, 0.7} 
\definecolor{tabthird}{rgb}{1, 1, 0.7} 

\definecolor{placeholder}{rgb}{0.6,0.8,0.95}
\newcommand{\placeholder}[1]{\textcolor{placeholder}
{\rule{\linewidth}{#1}}}

\newcommand{\picube}{\(\pi^3\)}
\newcommand{\mogedepth}{\widehat{d}}
\newcommand{\solveddepth}{d^*}
\newcommand{\depth}{d}
\newcommand{\gtdepth}{\depth}
\newcommand{\domain}{\mathrm{\Omega}}
\newcommand{\neighb}{\mathcal{N}}
\newcommand{\weight}{w}

\newcommand{\hanyu}[1]{\textcolor[HTML]{2BB673}{{[#1]}}}

\newcommand{\copyrightgoogle}{\href{https://www.google.com/help/terms_maps/}{\textcopyright Google}\xspace}

\titlerunning{\metricscenes}

\author{Yuanbo Xiangli\inst{1,2} \and
Hanyu Chen\inst{1} \and
Xueqing Tsang\inst{1}\and Noah Snavely\inst{1}}

\authorrunning{Y.~Xiangli et al.}

\institute{\inst{1} Cornell University, \quad \inst{2} Shanghai Jiao Tong University}

\maketitle

\vspace{-10pt}
\begin{abstract}
Metric scale monocular geometry estimation has seen significant progress through large-scale data aggregation, yet current foundation models suffer from a persistent ``scale-collapse'' phenomenon: distant landmarks and vast landscapes are metrically underestimated. 
This performance gap stems from a training data bottleneck, where existing metric-scale datasets are hardware-constrained to 
unvaried
street-level LiDAR or short-range indoor scans,
or consist of 
synthetic data that lacks the semantic complexity of the physical world. 
To bridge this gap, we curate a new metrically-grounded, in-the-wild 
dataset 
that we call \emph{\metricscenes}, gathered from a variety of sources including Internet photo collections and stereo imagery. 
We estimate camera poses and initial depth maps for each scene using off-the-shelf methods, and recover absolute scale from geo-tagged metadata as well as known stereo camera baselines.
We also improve the quality of depth maps derived from \metricscenes 
via a new two-stage Poisson completion method. 
Fine-tuning MoGe-2 on our dataset significantly mitigates scale-collapse and achieves superior metric accuracy in unconstrained, open-domain scenes while maintaining state-of-the-art performance on standard benchmarks.
Project page: \url{https://metricscenes.github.io/}.

\end{abstract}    
\section{Introduction}
\label{sec:intro}

\begin{figure}[t]
  \centering
    \includegraphics[width=.95\linewidth]{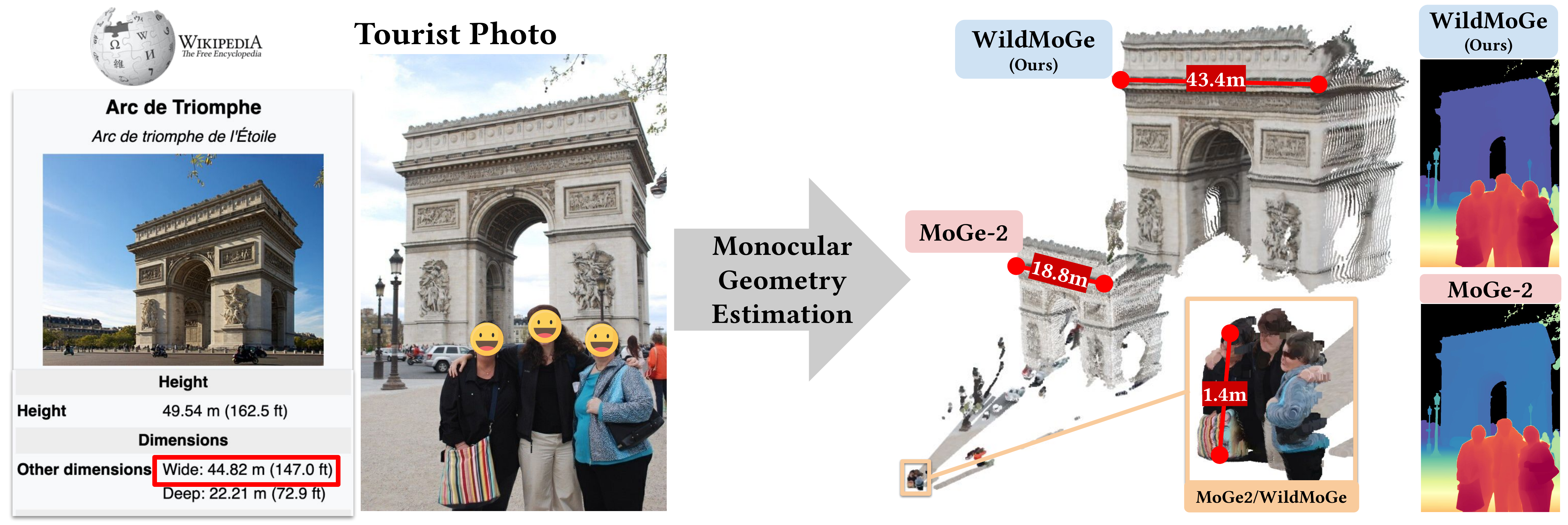}
    \vspace{-.75em}
   \caption{\small\textbf{Scale-collapse in metric geometry estimation.} State-of-the-art monocular geometry estimation (MGE) methods like MoGe-2~\cite{Wang2025MoGe2} exhibit a persistent scale-collapse phenomenon, where far-field structures are metrically shrunk and pulled toward the camera. According to Wikipedia, the real Arc de Triomphe has a physical width of $44.8$~m, but MoGe-2 predicts a metric point cloud of width just $18.8$~m. 
   By fine-tuning on our proposed MetricScenes dataset, our model, WildMoGe, recovers a more faithful landmark scale ($43.4$~m) while maintaining consistent metric estimation for foreground subjects ($\sim1.4$~m).}
   \label{fig:teaser}
   \vspace{-2em}
\end{figure}

Determining absolute scene scale from a single image is an inherently ill-posed task.
Consider the tourist photo of the Arc de Triomphe in Fig.~\ref{fig:teaser}.
From visual information alone, we can't be completely certain that we're looking at the monumental real-world landmark, as opposed to a  
cleverly constructed toy scene.
However, 
there are visual cues to absolute scale, some of which humans rely on, such as 
reference objects of known scale~(\eg, people).
Modern monocular geometry estimation (MGE) models ought to be able to perform a similar deduction via such semantic priors and other cues.

However, we found that current state-of-the-art methods frequently fail 
to estimate correct scene scale, leading to a persistent \emph{scale-collapse} phenomenon in ``in-the-wild'' scenarios. 
Fig.~\ref{fig:teaser} shows an example where clear semantic references (people) are present, yet where models like MoGe-2~\cite{Wang2025MoGe2} exhibit a significant scale inconsistency across the range of distances: 
the predicted metric scale for near-field objects is plausible---in this case, the tourists have a plausible height---yet the scale for far-field structures is dramatically underestimated---here, the Arc de Triomphe in the background is metrically predicted to be just $18.8$~m wide, which is more than 2\(\times\) smaller than the ground truth width ($44.8$~m).
MoGe-2 has posited a miniaturized landmark, despite cues to the contrary.

We argue that this performance gap is 
due to 
limitations in available training data. 
Current metric-scale, real-world datasets are restricted to two narrow categories due to hardware constraints: 
1) terrestrial LiDAR datasets like KITTI~\cite{Uhrig2017kitti} and Argoverse 2~\cite{Wilson2023Argoverse2N}, which are predominantly confined to vehicle-mounted street scenarios featuring limited diversity, and
2) active 3D scanner-captured data, such as ScanNet++~\cite{Yeshwanth2023ScanNetAH} and ARKitScenes~\cite{Baruch2021ARKitScenesA}, which 
are restricted by their short effective range to small-scale environments like indoor rooms.
Synthetic datasets attempt to bridge this gap through diverse perspectives and motions~\cite{Fonder2019MidAirAM, Wang2020TartanAirAD, Butler2012sintel},
but fail to capture the true metric diversity of the real world due to the reliance on procedural rules and limited 3D assets.

To enrich the availability of real-world metric data, we introduce \emph{\metricscenes}, a large-scale dataset curated from a variety of commonplace visual sources including Internet photo collections and stereo imagery. These sources provide the environmental and semantic variety missing from existing hardware-constrained datasets. We reconstruct camera viewpoints and initial depth maps via off-the-shelf methods, then recover absolute physical scale by leveraging geolocated landmark metadata and stereo camera baselines. Specifically, we aggregate data from AerialMegaDepth~\cite{vuong2025aerialmegadepth}, MegaScenes~\cite{Tung2024MegaScenesSV}, and Stereo4D~\cite{jin2025stereo4d}, and develop pipelines to extract metric-scale depth maps in each case.
We observe that depth maps derived from SfM and MVS often lack foreground transients. Naive Poisson completion fails here because the gradient guidance produced by monocular models also suffer from \emph{scale-collapse}, causing the solver to unnaturally enlarge foreground objects or distort boundaries. To remedy this, we propose a two-stage edge-aware Poisson completion algorithm that decouples background metric alignment from foreground integration, ensuring sharp boundaries and consistent metric scale throughout the view.
We show that fine-tuning MoGe-2 on \metricscenes significantly improves metric accuracy for unconstrained, in-the-wild photos. Our model generalizes effectively to novel scenes while maintaining performance comparable to state-of-the-art methods on standard benchmarks.
\section{Related Work}
\label{sec:related}
\noindent\textbf{Scale Ambiguity.}
Recovering absolute dimensions from a single image involves inherent scale ambiguity. Despite advances, resolving true physical scale remains a central challenge for monocular metric depth and geometry estimation (MDE/MGE).
Early works like ZoeDepth~\cite{bhat2023zoedepth} and LeRes~\cite{Yin2020LearningTR} attempted to bridge this gap using domain-specific metric heads or focal length decoupling. 
Recent foundation models have introduced more sophisticated alignment strategies.
The Metric3D series~\cite{Yin2023Metric3DTZ,Hu2024Metric3Dv2} transform images into a virtual camera space to resolve metric ambiguity across diverse data sources. 
UniDepth v2~\cite{piccinelli2025unidepthv2universalmonocularmetric} employs a self-prompting camera module, which simultaneously learns to predict camera embeddings.
MoGe-2~\cite{Wang2025MoGe2} decouples structural geometry from global scale, utilizing DINOv2 global tokens for scale prediction.

However, these architectural innovations are often insufficient for robust ``in-the-wild'' generalization. As shown in Fig.~\ref{fig:teaser}, state-of-the-art models suffer from scale-collapse in unconstrained environments, often underestimating the dimensions of distant structures. This performance gap suggests that the primary bottleneck is no longer architectural, but rather a fundamental lack of scale-diverse, real-world training data.

\medskip\noindent\textbf{Metric-scale Datasets.}
Real-world metric depth datasets relied on physical sensors like terrestrial LiDAR~(\eg, KITTI~\cite{Uhrig2017kitti}, Argoverse 2~\cite{Wilson2023Argoverse2N}, Waymo~\cite{Sun2019ScalabilityIP}) and active 3D scanners~(\eg, ScanNet++~\cite{Yeshwanth2023ScanNetAH}, ARKitScenes~\cite{Baruch2021ARKitScenesA}). While scale-accurate, these are restricted to driving corridors with limited diversity or small-scale indoor rooms and often suffer from spatial misalignment and sensor asynchrony. To bypass these hardware limits, synthetic 
datasets, like urban-focused MatrixCity~\cite{Li2023MatrixCityAL} and Synscapes~\cite{Wrenninge2018SynscapesAP}, indoor-oriented Hypersim~\cite{Roberts2020HypersimAP}, and aerial-view TartanAir~\cite{Wang2020TartanAirAD}, provide noise-free, dense labels across diverse domains. 
To further improve generalization, internet-scale collections like MegaDepth~\cite{MegaDepthLi18} and BlendedMVS~\cite{Yao2019BlendedMVSAL} use structure from motion methods~\cite{schoenberger2016sfm} to reconstruct scenes with wide semantic diversity, though they recover geometry only up to an unknown scale.

We analyzed standard training data and found a critical imbalance: large-scale in-the-wild SfM datasets offer semantic variety but lack metric scale. Metric supervision is currently split between LiDAR-based datasets, comprising 59\% of outdoor metric frames but limited to driving scenarios; and the rest are synthetic data, which lacks real-world complexity. 
No existing large-scale real-world dataset simultaneously provides absolute scale and in-the-wild diversity. 
Our work addresses this gap by creating \metricscenes, a real-world metric-scale
dataset that 
provides depth training data that is both metric and sharp.

\section{Generating Metric-Scale 3D Data}
\label{sec:dataset_construction}
To fill critical gaps and add diversity missing in existing real-world metric datasets,
we leverage widely available visual sources, including Internet photo collections and stereo imagery. 
In Sec.~\ref{subsec:internet_photo_collection}, we describe 
how we process noisy Internet photo collections and recover physical scale using geo-tagged metadata. 
In Sec.~\ref{subsec:stereo_pair}, we cover how we acquire high-quality dense depth maps from stereo video sequences and recover metric scale using known camera baselines. 
The initial depth maps we obtain from off-the-shelf vision methods often contain holes and missing regions.
In Sec.~\ref{subsec:poisson_reconstruction}, we describe a new two-stage edge-aware depth completion strategy that preserves correct scene scale and geometry while enhancing 
depth sharpness.

\subsection{Internet Photo Collections}
\label{subsec:internet_photo_collection}
We gather imagery from AerialMegaDepth\cite{vuong2025aerialmegadepth} and MegaScenes~\cite{Tung2024MegaScenesSV}, which consist of diverse Internet-sourced photography, including historical archives, tourist snapshots, and professional imagery. Specifically,
AerialMegaDepth enhances MegaDepth~\cite{MegaDepthLi18} by jointly reconstructing landmarks with geo-tagged aerial views rendered from Google Earth. 
MegaScenes provides a vast collection of SfM reconstructions recovered at an arbitrary scale. 
We leverage geo-tagged images from online mapping services to anchor these reconstructions to absolute physical dimensions.

\begin{figure}[t]
    \centering
    \includegraphics[width=0.95\linewidth]{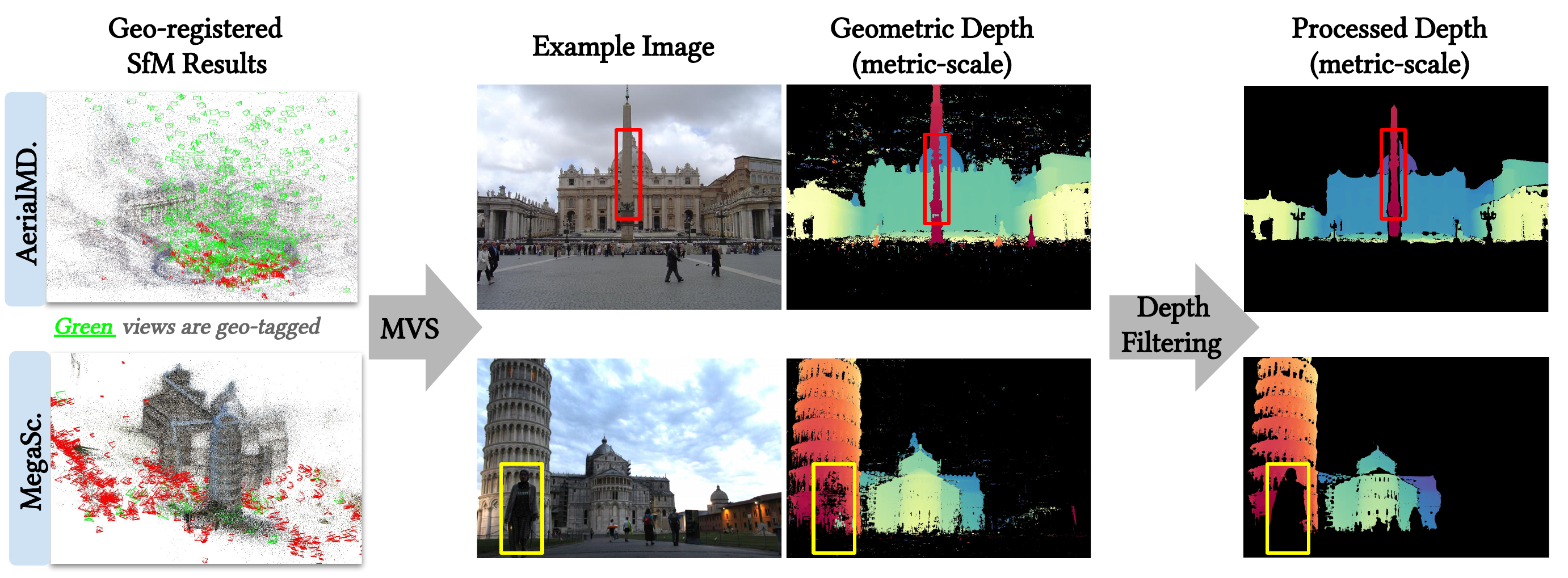}
    \vspace{-0.75em}
    \caption{\small\textbf{Metric depth from Internet photo collections.} \textcolor[HTML]{16d510}{Geo-tagged} images obtained from online mapping sites can be used to scale SfM results to absolute metric scale. AerialMegaDepth~\cite{vuong2025aerialmegadepth} is reconstructed with pseudo-synthetic views rendered from Google Earth and scenes are scaled accordingly. MegaScenes~\cite{Tung2024MegaScenesSV} contains natively unscaled SfM results. 
    We augment these SfM models with georeferenced street-level views to scale the geometry to physical dimensions. 
    After scaling and running MVS, we apply a depth filtering method to remove transient objects~(\textcolor[HTML]{f7d31a}{yellow} box) and filter out depth-bleeding regions~(\textcolor{red}{red} box).}
    \label{fig:amd_megas_pipeline}
    \vspace{-2em}
\end{figure}

\medskip\noindent\textbf{Robust SfM and MVS.}
For AerialMegaDepth, we use the SfM results provided in their dataset. For MegaScenes, we implement a pipeline addressing both structural errors and local depth inaccuracies. First, to resolve potential Doppelganger issues~\cite{cai2023doppelgangers}, where visually similar but geographically distant 3D surfaces cause spurious correspondences that mislead the reconstruction, we follow~\cite{xiangli2025doppelgangersimprovedvisualdisambiguation} to obtain sparse reconstruction results with MASt3R-SfM~\cite{Duisterhof2024MASt3RSfMAF} and the Doppelganger classifier.
For both AerialMegaDepth and MegaScenes,
after MVS~\cite{schoenberger2016mvs}, on the obtained geometric depths,
we apply a stability filtering strategy to identify and remove unstable pixels with high depth variance, and leverage MoGe-2's predictions to filter out depth-bleeding regions where background depth eats away foreground depth~\cite{MegaDepthLi18,li2026longtail}. This depth refinement pipeline ensures the final labels are geometrically consistent and free from dynamic artifacts. Examples are shown in Fig.~\ref{fig:amd_megas_pipeline}.

\medskip\noindent\textbf{Metric-Scale Recovery from Geolocations.}
To establish a globally consistent metric coordinate system, we leverage geo-referenced imagery to anchor our 3D reconstructions, as illustrated in Fig.~\ref{fig:amd_megas_pipeline}. AerialMegaDepth achieves metric scale by jointly reconstructing Internet photos with Google Earth renderings from specified, geolocated viewpoints.
Since the rendered viewpoints are inherently metrically grounded, the resulting models inherit the absolute metric scale.
Since MegaScenes lacks metric information, we leverage online mapping sites
to obtain geo-tagged street view images:
\begin{itemize}
    \item \textbf{Viewpoint sampling:} We ascertain the landmark's outline and sample approximately $100$ evenly distributed viewpoints along surrounding roads.
    \item \textbf{Coverage optimization:} To ensure full-facade coverage, we query for images with camera headings directed toward the scene center and apply geometric occlusion checks to remove obstructed views.
    \item \textbf{Data preference:} We prioritize imagery and metadata contributed by professional providers over those from individual users, as the former offers more accurate geolocation.
\end{itemize}
To register these geo-tagged images to existing reconstructed SfM models in MegaScenes, we use a method based on MASt3R. 
Specifically, MASt3R descriptors are used to compute matches between geo-tagged images and the existing reconstructed models. 
To resolve matching ambiguities arising from scene symmetries or repetitive elements, we use the Doppelganger classifier~\cite{xiangli2025doppelgangersimprovedvisualdisambiguation} to prune spurious correspondences in the scene graph. 
After registering the geo-tagged images, we use the COLMAP's model aligner~\cite{schoenberger2016sfm} to rotate the reconstruction into a Manhattan World frame, ensuring the dominant vertical and horizontal structures in the scenes are axis-aligned. 
To establish absolute metric scale and positioning, we apply RANSAC to robustly estimate a similarity transformation between the recovered 3D camera positions and their corresponding Earth-Centered, Earth-Fixed (ECEF) coordinates.
To ensure robustness, we discard scenes with implausible scale factors, high mean registration errors, or low RANSAC inlier ratios.

\subsection{Stereo Imagery}
\label{subsec:stereo_pair}
The Stereo4D dataset~\cite{jin2025stereo4d} provides a large-scale collection of diverse, real-world stereoscopic video sequences captured using VR180 devices.
The dataset originally offers roughly-estimated metric scaled depth maps, 
using the assumption that most VR180 stereo cameras have a baseline of approximately $6.3$~cm (optimizing for the average human interpupillary distance).
However, we observed that in practice, the physical baseline varies between different camera rigs. 
To ensure precise metric recovery, we utilize a subset of Stereo4D videos where camera configurations are explicitly documented in the video description.

\begin{figure}[t]
    \centering
    \includegraphics[width=.95\linewidth]{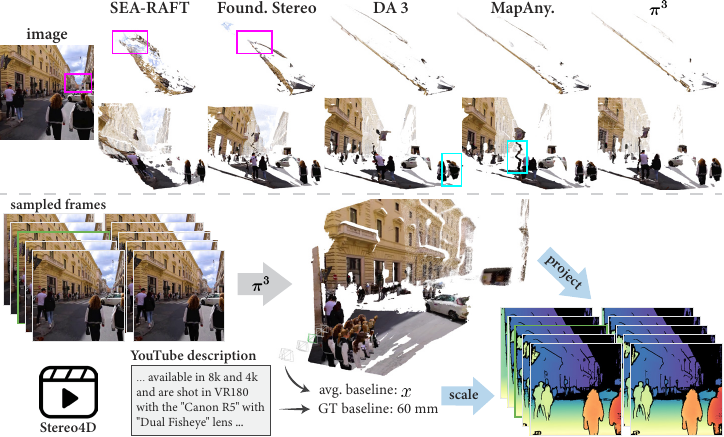}
    \vspace{-0.7em}
    \caption{\small\textbf{Metric Depth from Stereo4D~\cite{jin2025stereo4d}.} Top: Standard stereo matching~\cite{wang2024sea, wen2025foundationstereo} often produces distorted geometry in poorly calibrated in-the-wild videos, as seen in the converging facades (\textcolor[HTML]{ff00ff}{magenta} boxes). Among multi-view models~\cite{wang2025pi,Lin2025DepthA3,keetha2026mapanything}, \picube{}~\cite{wang2025pi} maintains the most robust geometry and sharp local details (\textcolor[HTML]{00ffff}{cyan} boxes). Bottom: We process stereoscopic sequences via \picube{} to obtain dense geometry and poses, then compute a global scale factor to align the predicted baseline with the camera's physical baseline. This yields accurately scaled metric depth maps.}
    \label{fig:pi3_pipeline}
   \vspace{-2em}
\end{figure}

\medskip\noindent\textbf{Robust Multi-View Reconstruction.}
Stereo4D originally relies on an optical flow estimator, SEA-RAFT~\cite{wang2024sea}, to derive scene geometry. 
However, we find that imperfect stereo calibration often causes both optical flow and dedicated stereo matching algorithms (e.g., FoundationStereo~\cite{wen2025foundationstereo}) to produce unreliable results. 
As illustrated in Fig.~\ref{fig:pi3_pipeline}, the recovered geometry exhibits severe surface distortion, where parallel building facades unnaturally converge toward a distant point.

Therefore, to generate more reliable depth maps, we propose to use a \emph{multi-view reconstruction} pipeline that re-estimates camera calibration, poses, and depth.
For each video clip in Stereo4D, we first sample a target center frame and extract \(N=16\) surrounding frames with a temporal stride of \(K=3\).
This spacing ensures sufficient temporal context and prevents the inclusion of many near-duplicate frames. 
Rather than relying on per-frame two-view stereo, we pass all $N+1$ stereo pairs to a multi-view reconstruction model to jointly estimate camera poses and scene geometry. 
This approach leverages broader spatio-temporal context, leading to more accurate and consistent depth estimates than single-pair methods.
We evaluated several methods, including \picube~\cite{wang2025pi}, DepthAnything V3~\cite{Lin2025DepthA3} and MapAnything~\cite{keetha2026mapanything}, and ultimately selected \picube{} due to its geometric robustness and ability to recover sharp local details (see Fig.~\ref{fig:pi3_pipeline}).

\medskip\noindent\textbf{Metric-Scale Recovery from Camera Baselines.}
Since \picube{} produces reconstructions in an arbitrary coordinate frame, we recover the absolute metric scale for each scene by aligning the average predicted baseline between the reconstructed stereo pairs to the known physical baseline. Specifically, denoting the physical baseline of the stereo rig as \(b_\text{gt}\), we compute a global scale factor \(s\) as:
\begin{equation}
    s = \frac{b_\text{gt}}{\frac{1}{N+1}\sum_{i=0}^{N} \|\mathbf{t}_\text{L}^{(i)} - \mathbf{t}_\text{R}^{(i)}\|_2},
    \label{eq:scale_factor}
\end{equation}
where \(\mathbf{t}_\text{L}^{(i)}\) and \(\mathbf{t}_\text{R}^{(i)}\) are the estimated translation vectors of the left and right cameras for the \(i\)-th sampled frame. Raw depth values \(\depth_{\text{pred}}\) are then scaled to produce the final metric depths: \(\depth_{\text{metric}} = s \cdot \depth_{\text{pred}}\).

\medskip\noindent\textbf{Frame Sampling and Filtering.}
Since frames within a clip are highly similar, we sample one sequence per clip and include only a single frame from the sequence in the final dataset to maximize dataset diversity. Specifically, we choose the \emph{center-left frame} in the sequence as it benefits from the most bidirectional temporal context during the multi-view reconstruction process.
To ensure high quality training data, we filter the target frame according to several criteria:
\begin{itemize}
    \item \textbf{Image quality:} We discard underexposed frames based on their median intensity. Additionally, we filter out motion-blurred frames that lack sufficient high-frequency detail based on their Tenengrad gradient energy.
    \item \textbf{Geometric consistency:} To ensure geometric accuracy, we perform a depth reprojection check on the stereo pair and discard the frame if more than 10\% of valid pixels are depth-inconsistent between the two views.
    \item \textbf{Calibration accuracy:} We reject frames violating expected stereo constraints. A frame is discarded if the inferred baseline deviates from the physical baseline by more than 10~mm (after scaling via Eq.~\ref{eq:scale_factor}) or if the relative camera rotation exceeds $1^\circ$.
\end{itemize}

To ensure reliable supervision from these pseudo-labels, we compute a robust scale factor between the derived metric depth from previous steps and the MoGe-2 prediction. If this factor exceeds $2\times$ or falls below $0.5\times$, we discard the frame as the scale is likely unreliable. This filtering is justified because \picube consistently recovers near-field geometry~(\eg pedestrians close to camera, nearby buildings as in Fig.~\ref{fig:pi3_pipeline}), a domain where MoGe-2 typically demonstrates high reliability.

\subsection{Depth Fusion and Completion}
\label{subsec:poisson_reconstruction}
Depth maps from prior stages are typically incomplete: Internet photos lack foregrounds and transient objects (Fig.~\ref{fig:amd_megas_pipeline}), while stereo imagery lacks backgrounds due to low confidence on distant points (Fig.~\ref{fig:pi3_pipeline}). Conversely, monocular models predict dense, continuous geometry but have problematic metric scale estimation.
As a result, these sources become complementary: our processed depth maps provide sparse metric anchors for scale, while monocular predictions offer structural guidance to fill holes and refine edges. Fusing them should produce depth maps that are both metrically accurate and visually complete.

\begin{figure}[t]
    \centering
    \includegraphics[width=0.9\linewidth]{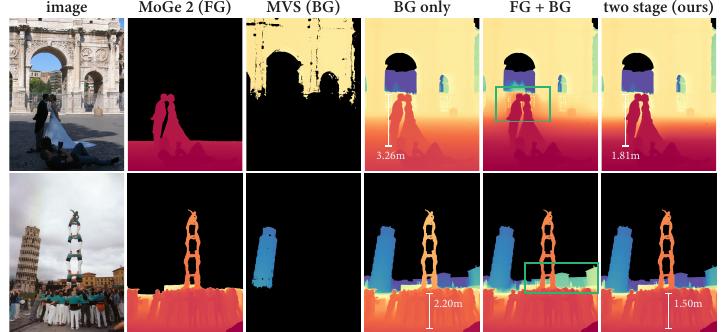}
    \vspace{-0.5em}
    \caption{\small\textbf{Visual comparison of depth completion methods.} With gradient guidance from MoGe-2, using only background anchors (BG only) preserves structure but uniformly mis-scales the entire map, yielding an implausible foreground (e.g., a $3.26$m tall person). Combining foreground and background anchors in a single stage (FG+BG) forces the solver to reconcile conflicting scales, causing depth bleeding and scale drift (\textcolor[HTML]{2BB673}{green} boxes). Our two-stage approach maintains accurate metric scale for both foreground and background while ensuring sharp, artifact-free boundaries.}
    \label{fig:poisson_comparison}
   \vspace{-2em}
\end{figure}

\medskip\noindent\textbf{Single-Pass Geometry Completion.} 
To reconstruct missing depth, MoGe-2 introduced a \emph{logarithmic-space Poisson completion} strategy. Their original approach integrates the incomplete ground truth depth map with guidance from a monocular depth model trained exclusively on synthetic data. Specifically, this is achieved by minimizing the gradient differences in the log domain:
\begin{equation}
    \min \sum_{i \in \domain} \|\nabla(\log \;\solveddepth_i) - \nabla(\log \;\mogedepth_i)\|^2, \quad \text{s.t.} \quad \solveddepth_i = \gtdepth_i, \;\forall i \in \partial \domain,
    \label{eq:geometry_completion}
\end{equation}
where \(\solveddepth_i\) is the solved depth, \(\mogedepth_i\) is the predicted depth providing gradient guidance, and \(\gtdepth_i\) represent fixed ground truth anchors bordering missing regions.

We found that this strategy effectively completes Stereo4D depth maps but faces a critical limitation on AerialMegaDepth and MegaScenes due to the scale-collapse phenomenon. 
Specifically, MoGe-2 often predicts an incorrect scale for distant landmarks, while maintaining a plausible scale for foreground subjects (Fig.~\ref{fig:teaser}). 
Because the Poisson solver treats the depth map as a continuous field, a scale correction forced by background anchors propagates globally through these gradient constraints. 
This results in a proportionate expansion of foreground objects,~\eg the physically impossible 3.26~m tall person in Fig.~\ref{fig:poisson_comparison} (BG only), making foreground scale unreliable.

To resolve this scale conflict, we propose anchoring the Poisson completion by trusting MoGe-2 to recover the metric scale of foreground content while relying on filtered MVS results for the absolute scale of background regions.
A naive solution would be to adapt Eq.~\ref{eq:geometry_completion} to use both the MVS background and MoGe-2 foreground as fixed anchors. However, this approach fails because the solver is still forced to reconcile conflicting scales across the entire scene. As shown in Fig.~\ref{fig:poisson_comparison} (FG+BG), unanchored transition regions such as a building’s base are inevitably warped toward the foreground to satisfy global gradient constraints.

\begin{figure}[t]
    \centering
    \includegraphics[width=.95\linewidth]{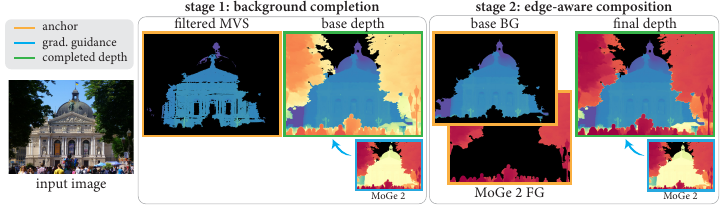}
    \vspace{-.75em}
    \caption{\small\textbf{Overview of the two-stage depth completion pipeline.} Stage 1 performs a Poisson solve using the gradient guidance of MoGe-2's depth map, constrained by the filtered MVS results. This yields a base depth map with an accurate background scale, though the foreground exhibits scale-drift. In Stage 2, the background isolated from the base depth map (base BG) and the MoGe-2 foreground (MoGe-2 FG) serve as joint anchors for the final completion. We perform an edge-weighted Poisson solve using these anchors, reusing MoGe-2's gradient guidance to produce a globally consistent metric depth map with sharp local details.}
    \label{fig:poisson_pipeline}
   \vspace{-2em}
\end{figure}

\medskip\noindent\textbf{Two-stage Edge-Aware Completion.} 
We attribute geometric distortion in single-stage solves to anchor sparsity, which provides insufficient constraints to override the scale-collapse inherent in the gradients of MoGe-2's depth prediction. Without a dense reference, the solver cannot reconcile the sparse metric anchors with a gradient field that pulls the background forward. This forces transition regions to warp as the geometry `stretches' to satisfy collapsed gradients between the fixed foreground and background metric anchors.
To address this, we propose a two-stage completion pipeline:
\begin{itemize}
    \item \textbf{Stage 1: Background Completion.} We first apply Eq.~\ref{eq:geometry_completion} using only sparse background depth as anchors, yielding results like in Fig.~\ref{fig:poisson_comparison} (BG only). In this stage, the background is reconstructed at a consistent metric scale without being warped by foreground depth constraints. We then discard the incorrectly scaled foreground regions, retaining only the completed background for the subsequent stage.
    \item \textbf{Stage 2: Edge-Aware Composition.} We combine the completed background from Stage 1 with the MoGe-2 predicted foreground to form a composite anchor. To preserve sharp boundaries and prevent depth-bleeding artifacts, the second Poisson solve utilizes an \emph{edge-weighted} objective:
    \begin{equation}
        \min \sum_{i \in \domain} \sum_{j \in \neighb(i)} \weight_{ij} \| (\log \;\solveddepth_i - \log \;\solveddepth_j) - (\log \;\mogedepth_i - \log \;\mogedepth_j) \|^2,
        \label{eq:two_stage}
    \end{equation}
    where the weight \(\weight_{ij}\) approaches zero across sharp depth discontinuities.
\end{itemize}

Example results are shown in Fig.~\ref{fig:poisson_comparison} (last column). Because the background is now densely anchored by the Stage 1 output, it remains fixed to its correct metric scale, while the edge weights ensure the foreground is integrated with crisp, artifact-free boundaries. 
Our full pipeline is illustrated in Fig.~\ref{fig:poisson_pipeline}.

\vspace{-1em}
\section{Evaluation}
\label{sec:evaluation}

\subsection{Dataset Overview}
The \metricscenes dataset contains $47{,}579$ images from $134$ scenes in Aerial-MegaDepth (only real-world images), $29,583$ images across $356$ scenes from MegaScenes, and $22{,}549$ frames from $1{,}725$ videos
\footnote{\scriptsize{There are $1{,}725$ videos, separated into $22{,}549$ clips, and we sample $1$ frame per clip.}} 
in Stereo4D. 
This diverse collection provides extensive coverage across a variety of perspectives and environments, including ground-level and aerial views, urban and natural landscapes, and both indoor and outdoor settings. 
We reserve a held-out set of $10$ scenes from AerialMegaDepth, $10$ scenes from MegaScenes, and $10$ videos from Stereo4D to serve as validation and test sets.

\subsection{WildMoGe}
We fine-tune the MoGe-2 ViT-Large-Normal model on our \metricscenes dataset for $10,000$ iterations with a batch size of $32$ (approximately $3$ epochs), ensuring model convergence while preventing significant divergence from the pre-trained geometric priors. We adopt the image cropping and data augmentation strategies from the original MoGe-2 implementation.
We use a learning rate of $1 \times 10^{-6}$ for the backbone and $1 \times 10^{-5}$ for the remaining parameters. Further details can be found in the supplemental material.

\begin{figure}[t]
    \centering
    \includegraphics[width=0.95\linewidth]{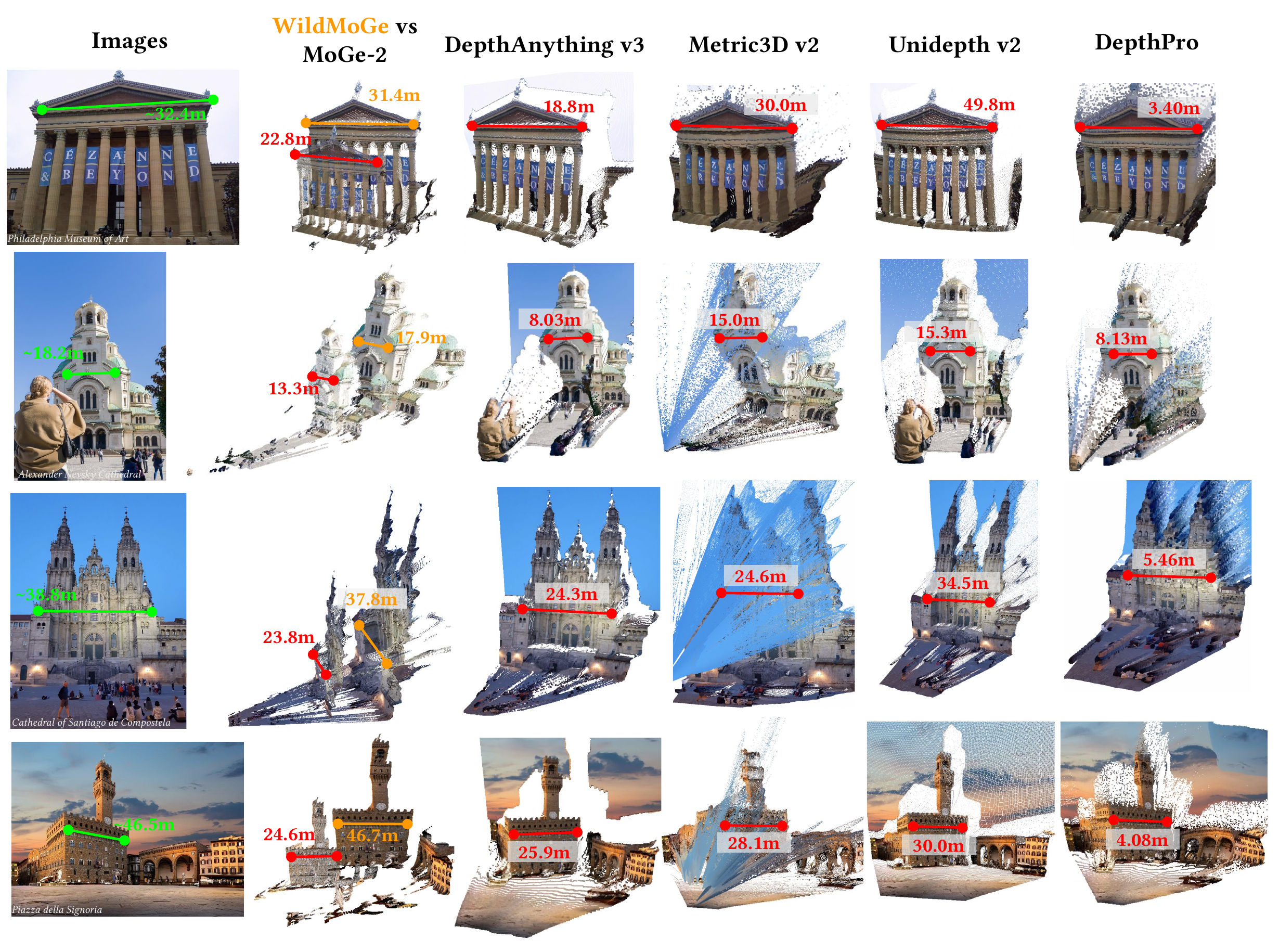}
    \vspace{-.75em}
    \caption{\small\textbf{Metrology of novel in-the-wild scenes.} The first column shows images with measurements obtained via Google Map's measuring tool. 
    We merge WildMoGe and MoGe-2's results into a single column to highlight the accurate scaling achieved by our training scheme. 
    WildMoGe consistently recovers more accurate absolute scales across diverse landmarks, whereas MoGe-2~\cite{Wang2025MoGe2}, DepthAnything v3~\cite{Lin2025DepthA3} and Metric3D v2~\cite{Hu2024Metric3Dv2} exhibit scale-collapse, underestimating the scale of background structures. While Unidepth v2~\cite{piccinelli2025unidepthv2universalmonocularmetric} produces more realistic scales, they still deviate from ground truth. DepthPro~\cite{depth_pro} often produces scales orders of magnitude smaller than reality. }
    \label{fig:qualitative_result}
    \vspace{-2em}
\end{figure}

\subsection{Qualitative Results}
\begin{figure}[t]
    \centering
    \includegraphics[width=0.95\linewidth]{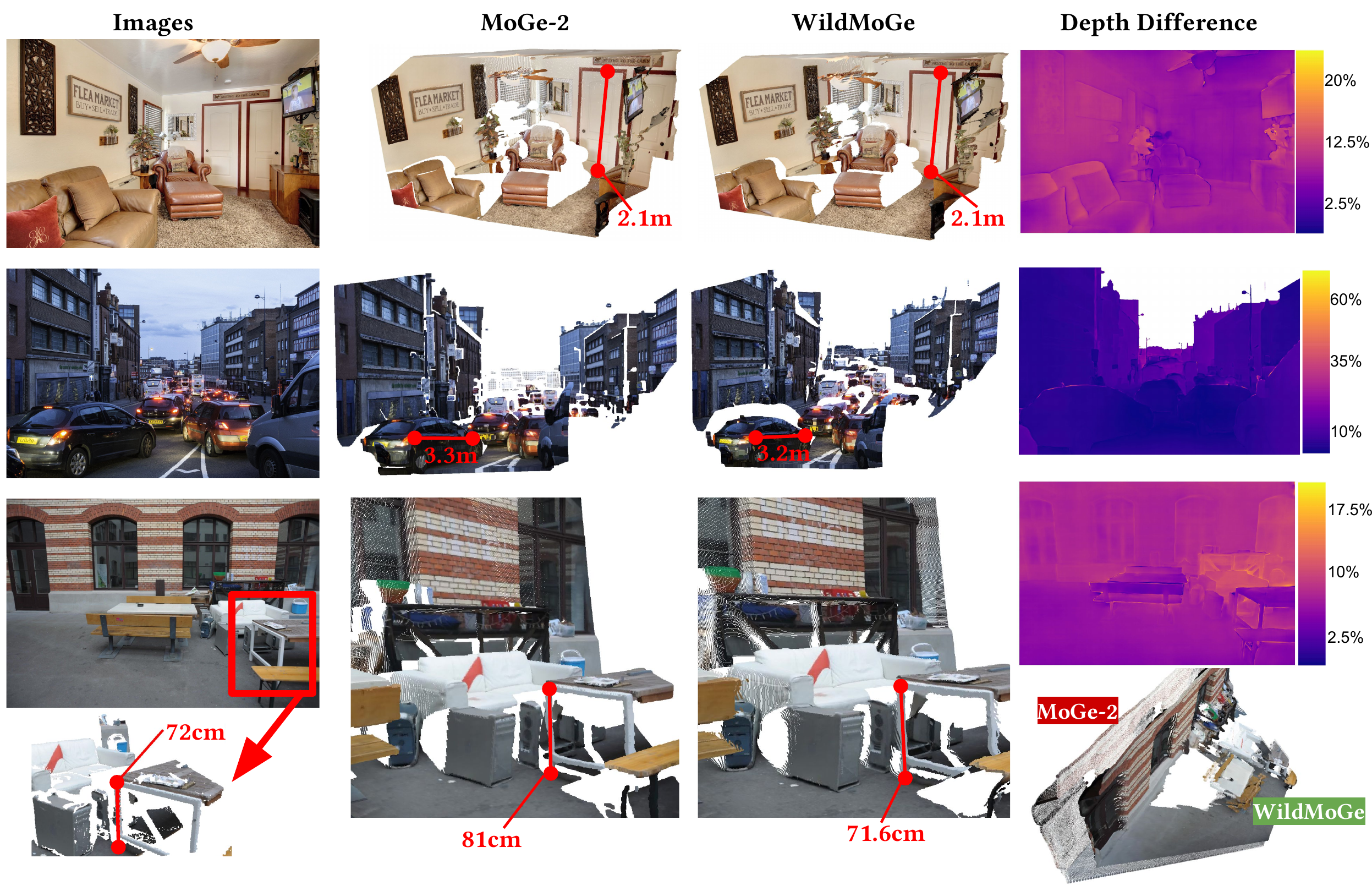}
    \vspace{-1em}
    \caption{\small\textbf{Comparison on the standard scenes.} We compare WildMoGe against MoGe-2~\cite{Wang2025MoGe2} on representative indoor and street-level scenes. In standard indoor and street contexts (Rows 1 \& 2), WildMoGe provides scale estimates consistent with MoGe-2. On the ETH3D~\cite{Schops2019ETH3D} courtyard scene (Row 3), WildMoGe achieves better accuracy, recovering a desk leg height of $71.6$cm compared to the $72$cm ground truth. This implies that WildMoGe's performance is driven by precise metric grounding rather than a bias toward larger scales.   }
    \label{fig:qualitative_standard}
    \vspace{-2em}
\end{figure}

In this section, we compare depth reconstructions of novel in-the-wild scenes obtained by our fine-tuned model, WildMoGe, as well as (vanilla) MoGe-2~\cite{Wang2025MoGe2}, DepthAnything v3~\cite{Lin2025DepthA3}, Metric3D v2~\cite{Hu2024Metric3Dv2}, UniDepth v2~\cite{piccinelli2025unidepthv2universalmonocularmetric}, and DepthPro~\cite{depth_pro}. These results are visualized in Fig.~\ref{fig:qualitative_result}. Additionally, we provide a comparison between WildMoGe and MoGe-2 on scenes prevalent in the standard training datasets, with results shown in Fig.~\ref{fig:qualitative_standard}.

From Fig.~\ref{fig:qualitative_result},
we observe that WildMoGe consistently recovers more accurate absolute scales across diverse landmarks, closely matching ground-truth dimensions~(\eg, $31.4$m vs. $32.4$m for the Philadelphia Museum of Art, $46.7$m vs $46.5$m for Piazza della Signoria). MoGe-2, DepthAnything v3 and Metric3D v2
exhibit scale-collapse behavior, consistently underestimating the size of far-field structures. 
UniDepth v2 produces more realistic scales but still deviates from ground truth,
and DepthPro often fails to recover absolute scale, producing results that are orders of magnitude smaller than reality. 
Note that these scenes are absent from the training set. This performance demonstrates that WildMoGe can generalize to unseen content, as opposed to simply memorizing training scenes.

Fig.~\ref{fig:qualitative_standard} demonstrates that WildMoGe provides scale estimates consistent with MoGe-2 in common indoor and street environments. For instance, both models estimate a door height of $2.1$m and a car length of approximately $3.2$m to $3.3$m. 
However, on the ETH3D~\cite{Schops2019ETH3D} courtyard scene, WildMoGe recovers a more accurate, smaller scale than MoGe-2. Specifically, WildMoGe estimates a desk leg height of $71.6$cm, closely matching the $72$cm ground truth, whereas MoGe-2 overestimates it at $81$cm. This indicates that WildMoGe is \emph{not} merely biased toward predicting larger scales as in Fig.~\ref{fig:teaser}\&~\ref{fig:qualitative_result}, but is instead performing reliable estimation grounded in absolute metric geometry. 

\subsection{Quantitative Evaluation}
Obtaining precise and dense ground-truth data for unconstrained, in-the-wild scenes is inherently challenging. 
Therefore, we provide quantitative results on our curated test set while acknowledging that these results may be less definitive due to potential imperfections in the test set labels. To provide a more comprehensive assessment, we also conduct evaluations on the standard benchmarks to demonstrate that our model is comparable to state-of-the-art performance in specialized environments while effectively bridging the gap between hardware-constrained training data and unconstrained, in-the-wild scenarios.

\medskip\noindent\textbf{Benchmark.}
For the \metricscenes test set, we treat the partial dense depth maps generated \emph{before} the Poisson completion step (Sec.~\ref{subsec:poisson_reconstruction}) as our ground truth. 
For Internet photo collections, these labels are filtered geometric depth maps from MVS that have been metrically scaled using geo-tagged metadata (Sec.~\ref{subsec:internet_photo_collection}).
For stereo imagery, the labels are from \picube{} multi-view reconstruction results scaled by camera baselines (Sec.~\ref{subsec:stereo_pair}).
We also follow MoGe-2~\cite{Wang2025MoGe2} to evaluate the accuracy of our model on $10$ standard datasets: NYUv2~\cite{Silberman2012nyuv2}, KITTI~\cite{Uhrig2017kitti}, ETH3D~\cite{Schops2019ETH3D}, iBims-1~\cite{ibim1_1,ibims1_2}, GSO~\cite{downs2022googlescannedobjects}, Sintel~\cite{Butler2012sintel}, DDAD~\cite{ddadpacking}, DIODE~\cite{diode_dataset}, Spring~\cite{Mehl2023Spring}, and HAMMER~\cite{jung2023hammer}. 

\medskip\noindent\textbf{Baselines.}
We compare our method to MoGe-2~\cite{Wang2025MoGe2}, UniDepth V2~\cite{piccinelli2025unidepthv2universalmonocularmetric}, Depth Pro~\cite{depth_pro},  MASt3R~\cite{leroy2024mast3r}, Depth Anything V2~\cite{Yang2024DepthAnythingV2} \& V3~\cite{Lin2025DepthA3}, ZoeDepth~\cite{bhat2023zoedepth} and Metric3D V2~\cite{Hu2024Metric3Dv2}.

\begin{table*}[t]
    \scriptsize
    \setlength{\tabcolsep}{1.2pt}
    \centering
    \caption{\small\textbf{Quantitative evaluation of relative and metric geometry.} The top section evaluates on the standard benchmarks, while the bottom section evaluates on \metricscenes test set. Metrics are color-coded: \colorbox[rgb]{0.94, 0.75, 0.77}{red} (best) and \colorbox[rgb]{0.98, 0.93, 0.75}{yellow} (second best).}
    \vspace{-1em}
    \resizebox{\textwidth}{!}{%
    \begin{tabular}{l|cc|cc|cc|cc|cc|cc|cc|cc|cc}
        \specialrule{.12em}{0em}{0em}
        
        \multirow{4}{*}{\textbf{Method}} 
        & \multicolumn{12}{c|}{\textbf{Relative Geometry}} 
        & \multicolumn{6}{c}{\textbf{Metric Geometry}} \\
        
        & \multicolumn{6}{c|}{Point} 
        & \multicolumn{6}{c|}{Depth} 
        & \multicolumn{2}{c|}{Point} 
        & \multicolumn{4}{c}{Depth} \\
        
        & \multicolumn{2}{c|}{\scriptsize Scale-inv.} 
        & \multicolumn{2}{c|}{\scriptsize Affine-inv.}
        & \multicolumn{2}{c|}{\scriptsize Local}
        & \multicolumn{2}{c|}{\scriptsize Scale-inv.}
        & \multicolumn{2}{c|}{\scriptsize Affine-inv.}
        & \multicolumn{2}{c|}{\scriptsize Affine-inv. \tiny(disp)}
        & \multicolumn{2}{c|}{\scriptsize (w/o GT Cam)}
        & \multicolumn{2}{c|}{\scriptsize (w/o GT Cam)}
        & \multicolumn{2}{c}{\scriptsize (w/ GT Cam)} \\
        
        &\scriptsize Rel\textsuperscript{p}\scriptsize$\downarrow$ &\scriptsize $\delta_1^\text{p}$\scriptsize$\uparrow$ 
        &\scriptsize Rel\textsuperscript{p}\scriptsize$\downarrow$ &\scriptsize $\delta_1^\text{p}$\scriptsize$\uparrow$ 
        &\scriptsize Rel\textsuperscript{p}\scriptsize$\downarrow$ &\scriptsize $\delta_1^\text{p}$\scriptsize$\uparrow$ 
        &\scriptsize Rel\textsuperscript{d}\scriptsize$\downarrow$ &\scriptsize $\delta_1^\text{d}$\scriptsize$\uparrow$ 
        &\scriptsize Rel\textsuperscript{d}\scriptsize$\downarrow$ &\scriptsize $\delta_1^\text{d}$\scriptsize$\uparrow$ 
        &\scriptsize Rel\textsuperscript{d}\scriptsize$\downarrow$ &\scriptsize $\delta_1^\text{d}$\scriptsize$\uparrow$ 
        &\scriptsize Rel\textsuperscript{p}\scriptsize$\downarrow$ &\scriptsize $\delta_1^\text{p}$\scriptsize$\uparrow$ 
        &\scriptsize Rel\textsuperscript{d}\scriptsize$\downarrow$ &\scriptsize $\delta_1^\text{d}$\scriptsize$\uparrow$ 
        &\scriptsize Rel\textsuperscript{d}\scriptsize$\downarrow$ &\scriptsize $\delta_1^\text{d}$\scriptsize$\uparrow$ \\
        \hline
        \multicolumn{19}{c}{\textit{Evaluation on standard benchmarks}} \\
        \hline

        ZoeDepth 
        & - & - & - & - & - & - 
        & 12.7 & 83.9 & 10.1 & 88.5 & 11.1 & 88.3 
        & - & - & 39.3 & 49.9 & - & - \\

        Metric3D V2
        & - & - & - & - & - & - 
        & 7.92 & \cellcolor[rgb]{0.98, 0.93, 0.75}91.8 & 7.66 & 92.9 & 9.51 & 89.4 
        & - & - & - & - & 18.3 & 73.9 \\
        
        DA V2 
        & - & - & - & - & - & - 
        & 10.7 & 87.6 & 8.48 & 90.8 & 8.82 & 91.6 
        & - & - & 29.9 & 56.6 & - & - \\ 

        DA V3 
        & 12.0 & 86.5 & 9.42 & 88.9 & 7.18 & 93.4 
        & 8.99 & 89.4 & 7.40 & 91.8 & 8.63 & 91.4 
        & 9.71 & 89.3 & 18.0 & 67.7 & - & - \\ 

        MASt3R
        & 14.5 & 82.1 & 11.6 & 86.0 & 8.09 & 92.2 
        & 11.2 & 86.5 & 9.38 & 89.1 & 11.6 & 87.8 
        & 26.2 & 55.3 & 49.7 & 30.3 & - & - \\
   
        UniDepth V2
        & 11.6 & 87.7 & 8.56 & 91.9 & 6.34 & 94.9 
        & 8.61 & 90.8 & 6.42 & 93.9 & 7.35 & 93.0 
        & 10.1 & 91.9 & 21.3 & \cellcolor[rgb]{0.98, 0.93, 0.75}75.3 & 18.5 & 82.6 \\
        
        Depth Pro
        & 12.4 & 87.7 & 9.93 & 89.4 & 6.91 & 94.1 
        & 9.81 & 89.1 & 7.65 & 92.0 & 8.42 & 91.7 
        & 13.7 & 81.9 & 27.6 & 54.4 & - & - \\

        MoGe-2
        & \cellcolor[rgb]{0.98, 0.93, 0.75}10.8 & \cellcolor[rgb]{0.98, 0.93, 0.75}88.5 & \cellcolor[rgb]{0.98, 0.93, 0.75}7.98 & \cellcolor[rgb]{0.98, 0.93, 0.75}91.7 & \cellcolor[rgb]{0.94, 0.75, 0.77}5.33 & \cellcolor[rgb]{0.94, 0.75, 0.77}95.9 
        & \cellcolor[rgb]{0.94, 0.75, 0.77}7.35 & \cellcolor[rgb]{0.94, 0.75, 0.77}92.2 & \cellcolor[rgb]{0.94, 0.75, 0.77}5.62 & \cellcolor[rgb]{0.94, 0.75, 0.77}94.8 & \cellcolor[rgb]{0.94, 0.75, 0.77}6.66 & \cellcolor[rgb]{0.94, 0.75, 0.77}93.8 
        & \cellcolor[rgb]{0.94, 0.75, 0.77}8.19 & \cellcolor[rgb]{0.94, 0.75, 0.77}93.6 & \cellcolor[rgb]{0.98, 0.93, 0.75}15.7 & \cellcolor[rgb]{0.94, 0.75, 0.77}76.8 & \cellcolor[rgb]{0.98, 0.93, 0.75}13.6 & \cellcolor[rgb]{0.94, 0.75, 0.77}87.4 \\

        \emph{WildMoGe}
        & \cellcolor[rgb]{0.94, 0.75, 0.77}10.1 & \cellcolor[rgb]{0.94, 0.75, 0.77}89.3 & \cellcolor[rgb]{0.94, 0.75, 0.77}7.68 & \cellcolor[rgb]{0.94, 0.75, 0.77}92.2 & \cellcolor[rgb]{0.98, 0.93, 0.75}5.50 & \cellcolor[rgb]{0.98, 0.93, 0.75}95.8 
        & \cellcolor[rgb]{0.98, 0.93, 0.75}7.49 & \cellcolor[rgb]{0.98, 0.93, 0.75}91.8 & \cellcolor[rgb]{0.98, 0.93, 0.75}5.63 & \cellcolor[rgb]{0.98, 0.93, 0.75}94.7 & \cellcolor[rgb]{0.98, 0.93, 0.75}6.97 & \cellcolor[rgb]{0.98, 0.93, 0.75}93.7 
        & \cellcolor[rgb]{0.98, 0.93, 0.75}8.39 & \cellcolor[rgb]{0.98, 0.93, 0.75}93.1 & \cellcolor[rgb]{0.94, 0.75, 0.77}15.6 & 73.4 & \cellcolor[rgb]{0.94, 0.75, 0.77}13.5 & \cellcolor[rgb]{0.98, 0.93, 0.75}84.9 \\
        \hline
        \multicolumn{19}{c}{\textit{Evaluation on \metricscenes test set}} \\
        \hline
        MoGe-2
        & \cellcolor[rgb]{0.98, 0.93, 0.75}6.76 & \cellcolor[rgb]{0.98, 0.93, 0.75}95.0
        & \cellcolor[rgb]{0.98, 0.93, 0.75}4.56 & \cellcolor[rgb]{0.98, 0.93, 0.75}96.9
        & - & -
        & \cellcolor[rgb]{0.98, 0.93, 0.75}4.85 & \cellcolor[rgb]{0.98, 0.93, 0.75}95.4
        & \cellcolor[rgb]{0.98, 0.93, 0.75}3.42 & \cellcolor[rgb]{0.98, 0.93, 0.75}97.4
        & \cellcolor[rgb]{0.98, 0.93, 0.75}4.61 & \cellcolor[rgb]{0.98, 0.93, 0.75}96.3
        & \cellcolor[rgb]{0.98, 0.93, 0.75}11.7 & \cellcolor[rgb]{0.98, 0.93, 0.75}87.3
        & \cellcolor[rgb]{0.98, 0.93, 0.75}37.2 & \cellcolor[rgb]{0.98, 0.93, 0.75}37.7
        & \cellcolor[rgb]{0.98, 0.93, 0.75}35.1 & \cellcolor[rgb]{0.98, 0.93, 0.75}44.0 \\

        WildMoGe
        & \cellcolor[rgb]{0.94, 0.75, 0.77}5.24 & \cellcolor[rgb]{0.94, 0.75, 0.77}97.0
        & \cellcolor[rgb]{0.94, 0.75, 0.77}3.67 & \cellcolor[rgb]{0.94, 0.75, 0.77}97.9
        & - & -
        & \cellcolor[rgb]{0.94, 0.75, 0.77}4.02 & \cellcolor[rgb]{0.94, 0.75, 0.77}97.1
        & \cellcolor[rgb]{0.94, 0.75, 0.77}2.98 & \cellcolor[rgb]{0.94, 0.75, 0.77}98.2
        & \cellcolor[rgb]{0.94, 0.75, 0.77}4.15 & \cellcolor[rgb]{0.94, 0.75, 0.77}97.1
        & \cellcolor[rgb]{0.94, 0.75, 0.77}7.59 & \cellcolor[rgb]{0.94, 0.75, 0.77}93.4
        & \cellcolor[rgb]{0.94, 0.75, 0.77}26.5 & \cellcolor[rgb]{0.94, 0.75, 0.77}73.8
        & \cellcolor[rgb]{0.94, 0.75, 0.77}26.0 & \cellcolor[rgb]{0.94, 0.75, 0.77}79.1\\

    \specialrule{.12em}{0em}{0em}
    \end{tabular}
    }
   \vspace{-1.75em}
    \label{tab:unified_geometry_evaluation}
\end{table*}

\medskip\noindent\textbf{Relative Geometry and Depth.}
Following MoGe-2, we evaluate relative geometry to assess how effectively each method reconstructs scene structure from a single input image. We measure performance across multiple representations, including scale-invariant and affine-invariant point maps, local point maps within each detected objects, and scale-invariant depth, as well as affine-invariant depth and disparity. Specifically, the relative error for point maps is defined as $Rel^{p} = \|\hat{\mathbf{p}}-\mathbf{p}\|_2 / \|\mathbf{p}\|_2$ and for depth maps as $Rel^{d} = |\hat{z}-z|/z$. We also report the percentage of inliers using the thresholds $\delta_1^p = (\|\hat{\mathbf{p}}-\mathbf{p}\|_2 / \|\mathbf{p}\|_2 < 0.25)$ and $\delta_1^d = \max(\hat{d}/d, d/\hat{d}) < 1.25$ across the 10 evaluation datasets. ZoeDepth, DepthAnything V2, and Metric3D V2 are excluded from the point-based evaluations as these models do not provide camera intrinsic predictions. On the \metricscenes test set, we omit local point map evaluation because our ground truth depth maps can be sparse.
Tab.~\ref{tab:unified_geometry_evaluation} (left side) presents quantitative results on relative geometry and depth prediction.
On the standard benchmarks, our WildMoGe achieves performance comparable to MoGe-2 while outperforming all other baselines. 
On the \metricscenes test set, WildMoGe consistently surpasses MoGe-2 across all metrics.
This indicates that fine-tuning on \metricscenes does not severely compromise the model's existing geometric priors; instead, it bridges the gap between near-field and far-field scale estimation on in-the-wild scenes, allowing the model to generalize effectively to unconstrained environments.

\medskip\noindent\textbf{Metric Geometry and Depth.}
We evaluate the accuracy of metric-scale geometry and depth across seven datasets with metric-scale annotations: NYUv2~\cite{Silberman2012nyuv2}, KITTI~\cite{Uhrig2017kitti}, ETH3D~\cite{Schops2019ETH3D}, iBims-1~\cite{ibim1_1,ibims1_2}, DDAD~\cite{ddadpacking}, DIODE~\cite{diode_dataset}, and HAMMER~\cite{jung2023hammer}. We report relative point error ($Rel^{p}$) and percentage of inliers ($\delta_1^{p}$) on predicted metric point maps. Similarly, we evaluate metric depth accuracy via relative depth error ($Rel^{d}$) and depth inlier percentage ($\delta_1^{d}$). Additionally, for methods that accept external camera parameters, we evaluate metric depth estimation using ground-truth camera intrinsics to isolate depth accuracy from errors in intrinsic parameter estimation.
Results are in Tab.~\ref{tab:unified_geometry_evaluation} (right side). 
While slightly worse than MoGe-2 on these metrics, WildMoGe outperforms most of the other baselines. 
This slight performance drop can be ascribed to the domain shift, and could be remedied by joint training.
On the \metricscenes test set, WildMoGe outperforms MoGe-2 across all metrics. This suggests that while there is a trade-off on legacy datasets due to domain shift, the model successfully overcomes the scale-collapse phenomenon to achieve superior accuracy in unconstrained, in-the-wild environments.

\begin{table*}[t]
    \scriptsize
    \setlength{\tabcolsep}{1.1pt}
    \centering
    \caption{\small \textbf{Ablation of finetuning MoGe-2 using different subsets of MetricScenes}. Evaluation is conducted on the standard benchmarks and \metricscenes's test set. We evaluate performance across relative geometry, metric geometry, and boundary sharpness. Metrics are color-coded: \colorbox[rgb]{0.94, 0.75, 0.77}{red} (best) and \colorbox[rgb]{0.98, 0.93, 0.75}{yellow} (second best). Boundary sharpness is reported on a subset of the standard benchmarks~\cite{Mehl2023Spring, Butler2012sintel, ibim1_1, jung2023hammer}, but is omitted for the \metricscenes test set.}
    \vspace{-1em}
    \resizebox{\textwidth}{!}{%
    \begin{tabular}{l|cc|cc|cc|cc|cc|cc|cc|cc|cc|c}
        \specialrule{.12em}{0em}{0em}
        
        \multirow{4}{*}{\textbf{Method}} 
        & \multicolumn{12}{c|}{\textbf{Relative Geometry}} 
        & \multicolumn{6}{c|}{\textbf{Metric Geometry}} 
        & \textbf{Sharpness} \\
        
        & \multicolumn{6}{c|}{Point} 
        & \multicolumn{6}{c|}{Depth} 
        & \multicolumn{2}{c|}{Point} 
        & \multicolumn{4}{c|}{Depth} 
        & Boundary \\
        
        & \multicolumn{2}{c|}{\scriptsize Scale-inv.} 
        & \multicolumn{2}{c|}{\scriptsize Affine-inv.}
        & \multicolumn{2}{c|}{\scriptsize Local}
        & \multicolumn{2}{c|}{\scriptsize Scale-inv.}
        & \multicolumn{2}{c|}{\scriptsize Affine-inv.}
        & \multicolumn{2}{c|}{\scriptsize Affine-inv. \tiny(disp)}
        & \multicolumn{2}{c|}{\scriptsize (w/o GT Cam)}
        & \multicolumn{2}{c|}{\scriptsize (w/o GT Cam)}
        & \multicolumn{2}{c|}{\scriptsize (w/ GT Cam)} 
        & \scriptsize  \\
        
        &\scriptsize Rel\textsuperscript{p}\scriptsize$\downarrow$ &\scriptsize $\delta_1^\text{p}$\scriptsize$\uparrow$ 
        &\scriptsize Rel\textsuperscript{p}\scriptsize$\downarrow$ &\scriptsize $\delta_1^\text{p}$\scriptsize$\uparrow$ 
        &\scriptsize Rel\textsuperscript{p}\scriptsize$\downarrow$ &\scriptsize $\delta_1^\text{p}$\scriptsize$\uparrow$ 
        &\scriptsize Rel\textsuperscript{d}\scriptsize$\downarrow$ &\scriptsize $\delta_1^\text{d}$\scriptsize$\uparrow$ 
        &\scriptsize Rel\textsuperscript{d}\scriptsize$\downarrow$ &\scriptsize $\delta_1^\text{d}$\scriptsize$\uparrow$ 
        &\scriptsize Rel\textsuperscript{d}\scriptsize$\downarrow$ &\scriptsize $\delta_1^\text{d}$\scriptsize$\uparrow$ 
        &\scriptsize Rel\textsuperscript{p}\scriptsize$\downarrow$ &\scriptsize $\delta_1^\text{p}$\scriptsize$\uparrow$ 
        &\scriptsize Rel\textsuperscript{d}\scriptsize$\downarrow$ &\scriptsize $\delta_1^\text{d}$\scriptsize$\uparrow$ 
        &\scriptsize Rel\textsuperscript{d}\scriptsize$\downarrow$ &\scriptsize $\delta_1^\text{d}$\scriptsize$\uparrow$ 
        &\scriptsize $F_1$$\uparrow$ \\
        \hline
        \multicolumn{20}{c}{\textit{Evaluated on the standard Benchmark}} \\
        \hline
        
        All
        & 10.1 & 89.3 & 7.68 & 92.2 & \cellcolor[rgb]{0.98,0.93,0.75}5.50 & \cellcolor[rgb]{0.94,0.75,0.77}95.8 
        & 7.49 & 91.8 & 5.63 & 94.7 & 6.97 & 93.7 
        & \cellcolor[rgb]{0.98,0.93,0.75}8.39 & \cellcolor[rgb]{0.98,0.93,0.75}93.1 & \cellcolor[rgb]{0.94,0.75,0.77}15.6 & 73.4 & \cellcolor[rgb]{0.98,0.93,0.75}13.5 & 84.9 & \cellcolor[rgb]{0.98,0.93,0.75}15.5 \\

        Stereo4D only
        & 11.3 & 88.0 & 8.00 & 91.8 & 5.52 & \cellcolor[rgb]{0.94,0.75,0.77}95.8 
        & 7.50 & \cellcolor[rgb]{0.98,0.93,0.75}92.3 & 5.65 & 94.8 & 6.77 & 93.7 
        & 9.29 & 91.4 & 19.0 & 63.4 & 15.0 & 82.2 & \cellcolor[rgb]{0.94,0.75,0.77}15.6 \\

        AerialMD. only
        & 9.81 & \cellcolor[rgb]{0.98,0.93,0.75}90.2 & 7.39 & \cellcolor[rgb]{0.94,0.75,0.77}92.9 & 5.59 & \cellcolor[rgb]{0.98,0.93,0.75}95.7 
        & \cellcolor[rgb]{0.98,0.93,0.75}7.41 & \cellcolor[rgb]{0.94,0.75,0.77}92.6 & 5.65 & \cellcolor[rgb]{0.98,0.93,0.75}94.9 & \cellcolor[rgb]{0.98,0.93,0.75}6.70 & \cellcolor[rgb]{0.98,0.93,0.75}93.8 
        & 8.60 & \cellcolor[rgb]{0.94,0.75,0.77}93.4 & 16.8 & \cellcolor[rgb]{0.94,0.75,0.77}78.9 & 14.6 & \cellcolor[rgb]{0.94,0.75,0.77}85.8 & 14.5 \\

        MegaSc. only
        & \cellcolor[rgb]{0.98,0.93,0.75}9.78 & \cellcolor[rgb]{0.98,0.93,0.75}90.2 & \cellcolor[rgb]{0.94,0.75,0.77}7.32 & \cellcolor[rgb]{0.98,0.93,0.75}92.8 & \cellcolor[rgb]{0.94,0.75,0.77}5.45 & \cellcolor[rgb]{0.94,0.75,0.77}95.8 
        & 7.45 & 91.9 & \cellcolor[rgb]{0.94,0.75,0.77}5.51 & \cellcolor[rgb]{0.94,0.75,0.77}95.0 & 6.75 & \cellcolor[rgb]{0.94,0.75,0.77}93.9 
        & 8.68 & 91.5 & \cellcolor[rgb]{0.98,0.93,0.75}15.9 & 73.0 & 13.7 & 82.5 & 15.0 \\

        AerialMD.$+$MegaSc.
        & \cellcolor[rgb]{0.94,0.75,0.77}9.73 & \cellcolor[rgb]{0.94,0.75,0.77}90.6 & \cellcolor[rgb]{0.98,0.93,0.75}7.33 & 92.1 & 5.52 & \cellcolor[rgb]{0.98,0.93,0.75}95.7 
        & \cellcolor[rgb]{0.94,0.75,0.77}7.40 & 92.1 & \cellcolor[rgb]{0.98,0.93,0.75}5.52 & \cellcolor[rgb]{0.98,0.93,0.75}94.9 & \cellcolor[rgb]{0.94,0.75,0.77}6.69 & \cellcolor[rgb]{0.94,0.75,0.77}93.9 
        & \cellcolor[rgb]{0.94,0.75,0.77}8.36 & 92.5 & 16.0 & \cellcolor[rgb]{0.98,0.93,0.75}74.2 & \cellcolor[rgb]{0.94,0.75,0.77}13.2 & \cellcolor[rgb]{0.98,0.93,0.75}85.1 & 14.9 \\

        \hline
        \multicolumn{20}{c}{\textit{Evaluated on \metricscenes Test Set}} \\
        \hline

        All
        & \cellcolor[rgb]{0.94,0.75,0.77}5.24 & \cellcolor[rgb]{0.94,0.75,0.77}97.0
        & \cellcolor[rgb]{0.94,0.75,0.77}3.67 & \cellcolor[rgb]{0.94,0.75,0.77}97.9
        &  - &  -
        & \cellcolor[rgb]{0.94,0.75,0.77}4.02 & \cellcolor[rgb]{0.94,0.75,0.77}97.1
        & \cellcolor[rgb]{0.94,0.75,0.77}2.98 & \cellcolor[rgb]{0.94,0.75,0.77}98.2
        & \cellcolor[rgb]{0.94,0.75,0.77}4.15 & \cellcolor[rgb]{0.94,0.75,0.77}97.1
        & \cellcolor[rgb]{0.94,0.75,0.77}7.59 & \cellcolor[rgb]{0.94,0.75,0.77}93.4
        & \cellcolor[rgb]{0.94,0.75,0.77}26.5 & \cellcolor[rgb]{0.94,0.75,0.77}73.8
        & \cellcolor[rgb]{0.94,0.75,0.77}26.0 & \cellcolor[rgb]{0.94,0.75,0.77}79.1 & -\\
        
        Stereo4D only
        & \cellcolor[rgb]{0.98,0.93,0.75}6.23 & \cellcolor[rgb]{0.98,0.93,0.75}96.6
        & \cellcolor[rgb]{0.98,0.93,0.75}3.83 & \cellcolor[rgb]{0.98,0.93,0.75}97.6
        &  - &  -
        & \cellcolor[rgb]{0.98,0.93,0.75}4.37 & \cellcolor[rgb]{0.98,0.93,0.75}97.0
        & \cellcolor[rgb]{0.98,0.93,0.75}3.04 & \cellcolor[rgb]{0.98,0.93,0.75}98.1
        & \cellcolor[rgb]{0.98,0.93,0.75}4.17 & \cellcolor[rgb]{0.98,0.93,0.75}97.0
        & 10.6 & 88.2
        & 34.0 & 39.3
        & 33.5 & 45.6 & - \\

        AerialMD. only
        & 6.33 & 95.6
        & 4.39 & 97.1
        &  - &  -
        & 4.55 & 96.0
        & 3.25 & 97.7
        & 4.48 & 96.4
        & 11.5 & 87.8
        & 41.0 & 58.7
        & 38.3 & 65.3 & - \\

        MegaSc. only
        & 6.68 & 94.5
        & 4.61 & 96.8
        &  - &  -
        & 5.13 & 94.9
        & 3.59 & 97.2
        & 4.82 & 96.0
        & 9.99 & 89.8
        & \cellcolor[rgb]{0.98,0.93,0.75}32.4 & 53.0
        & 28.7 & 61.9 & - \\
        
        AerialMD.$+$MegaSc.
        & 6.66 & 95.2
        & 4.52 & 97.0
        &  - &  -
        & 4.86 & 95.5
        & 3.43 & 97.5
        & 4.66 & 96.2
        & \cellcolor[rgb]{0.98,0.93,0.75}9.57 & \cellcolor[rgb]{0.98,0.93,0.75}91.0
        & 32.7 & \cellcolor[rgb]{0.98,0.93,0.75}65.0
        & \cellcolor[rgb]{0.98,0.93,0.75}26.4 & \cellcolor[rgb]{0.98,0.93,0.75}74.7 & - \\

    \specialrule{.12em}{0em}{0em}
    \end{tabular}
    }
   \vspace{-1.75em}
    \label{tab:ablation_metricscenes}
\end{table*}

\medskip\noindent\textbf{Fine-tuning on Different Subsets.}
The ablation results in Tab.~\ref{tab:ablation_metricscenes} validate the multi-source design of the \metricscenes dataset by showing how each subset contributes to performance.
Each ablated model is trained for $3$ epochs, consistent with the WildMoGe training schedule.
On the standard benchmarks, we observe that each data source addresses a specific aspect of the model's performance. Specifically, Stereo4D helps improve boundary sharpness ($F_1 = 15.6$); AerialMegaDepth serves as a reliable metric anchor, resulting in strong metric-scale geometry accuracy; MegaScenes adds important perspective diversity, and when combined with AerialMegaDepth, achieves the lowest scale-invariant point error, which helps the model handle many different camera viewpoints. Generally, we find that joint training on the full corpus leads to better metric geometry and sharper boundaries.
On the \metricscenes test set, we find that training exclusively on each subset tends to limit overall performance. In contrast, joint training across all subsets achieves superior results across every metric, suggesting a strong synergistic effect.

\section{Conclusion}
\label{sec:conclusion}
In this work, we identify scale collapse as a challenge in monocular metric depth/geometry estimation on in-the-wild scenes.
Despite the impressive performance of modern models on standard benchmarks, they often fail to maintain consistent real-world dimensions on certain types of data, such as distant landmarks or vast landscapes. 
We consider this fundamentally a data problem rooted in the limited scale diversity of existing training sets, 
which are typically restricted by hardware constraints or a lack of semantic complexity. 
To bridge this gap, we propose \metricscenes, a diverse and metrically-grounded real-world dataset curated from widely available sources including Internet photo collections and online stereo videos/imagery. 
Through qualitative and quantitative evaluations, we found that fine-tuning the state-of-the-art MoGe-2 model on our dataset effectively overcomes scale-collapse, enabling faithful metric scale recovery in unconstrained environments while maintaining competitive performance on standard benchmarks.

\section{Acknowledgment}
This work was supported by the Institute of Information \& Communications Technology Planning \& Evaluation (IITP) grant funded by the Korean Government (MSIT) (No. RS-2024-00457882, National AI Research Lab Project), and the Major Program of the National Natural Science Foundation of China (Grant No. 62595772).

%
%
\bibliographystyle{splncs04}
\bibliography{main}
\end{document}